# Motorcycle detection and classification in urban Scenarios using a model based on Faster R-CNN

**Jorge E. Espinosa[†], Sergio A. Velastin [††], and John W. Branch [†††]**

[†]Facultad de Ingenierías, Politécnico Colombiano Jaime Isaza Cadavid – Medellín –Colombia
[††] University Carlos III - Madrid Spain, Cortexica Vision Systems Ltd. UK and Queen Mary University of London, UK
[†††]Facultad de Minas, Universidad Nacional de Colombia – Sede Medellín
jeespinosa@elpoli.edu.co, sergio.velastin@theiet.org, jwbranch@unal.edu.co

## Abstract

This paper introduces a Deep Learning Convolutional Neutral Network model based on Faster-RCNN for motorcycle detection and classification on urban environments. The model is evaluated in occluded scenarios where more than 60% of the vehicles present a degree of occlusion. For training and evaluation, we introduce a new dataset of 7500 annotated images, captured under real traffic scenes, using a drone mounted camera. Several tests were carried out to design the network, achieving promising results of 75% in average precision (AP), even with the high number of occluded motorbikes, the low angle of capture and the moving camera. The model is also evaluated on low occlusions datasets, reaching results of up to 92% in AP.

## 1 Introduction

With an increasing number of motorcycles as a popular means of transport in emerging countries, there has been an important growth in accidentality and fatality rates. In Latin-America, fatal casualties involving motorbikes account for 45% of (all) traffic accidents [1]. Meanwhile, in other emerging regions as in the Middle East 63% of traffic accidents involve vulnerable road users (VRUs), of which 32% corresponds just to motorcyclists [2]. The World Health Organization (WHO) in 2015 established that the rate of VRUs involved in fatalities in traffic accidents represented more than 49% [3].

Therefore, it is important to implement techniques or strategies that allow detecting motorcycles for urban traffic analysis. Traditional urban traffic monitoring e.g. using inductive loop sensors, have limitations in discrimination capacity, maintenance and cost that makes video analysis an attractive alternative. However, video detection is a complex problem and high vehicle densities, illumination changes and even camera location (e.g. movement, displacement) can affect the final detection results. Traditionally, video techniques analysis requires reliable methods for object feature extraction to obtain accurate classification results. Most video detection systems are implemented building discrimination capabilities on appearance features or motion features. For motorcycle classification, appearance features include 3D models [4] [5], vehicle dimensions [6] [7] [8], symmetry, colour, shadow, geometrical features and texture and even wheel contours [9], as well as the use of stable features [10], HOG for evaluation of helmet presence [11] [12], Haar Like Features [13], Harris Corners [14], variations of HOG [15], and the use of SIFT, DSIFT and SURF [16]. Motion features are obtained based on traffic dynamics. Background subtraction is the main technique used [17], along with frame difference [18], Kalman filter [19], optical flow [20] [21], etc. A detailed survey of traditional vehicle detection methods is described in [22].

In the last eight years, a breakthrough has emerged in computer vision: *deep learning theory* (DL), specially applied to image processing. The use of Deep Learning has been already reported for vehicle detection. Pioneering approaches implement 2D Deep Belief Networks (2D-DBN) [23] to learn features, complemented with a pre-training sparse filtering process [24] or using Hybrid architectures (HDNN) to extract multi scale features. Some implementations use colour as a discriminative feature [25] [26]. To obtain real time implementations, pre-training schemes [27] are used even with low resolution images [28]. Based on object proposal algorithms, two stage CNN models integrate region proposal and classification in a single architecture, such as Fast R-CNN [29] and faster R-CNN [30] based models for vehicle detection and classification [31][32][33]. Motivated by safety measures, helmet detection in motorcycle riders has inspired research using geometrical features [34], hand crafted features (HOG, SIFT, LBP, CHT [35] [36] [12]), neuro-fuzzy detectors [37] and neural networks [38]. Nevertheless, there are few reports exploring CNNs for motorcycle classification, e.g. using a pre-trained network (AlexNet) for feature extraction as in [39] or as a motorcycle classifier and helmet detector [40].

This work uses a CNN model based on Faster R-CNN [30] for the task of detecting and classifying motorcycles in urban traffic video sequences, which are characterized by high occlusion (more than 60%). The paper is organized as follows: section 2 gives a brief explanation of CNNs along with an overview of Faster R-CNN networks and the advantage of its use for detection and classification. Section 3 introduces the annotated motorcycles dataset created for this research. Section 4 describes the proposed model inspired on Faster-RCNN. Section 5 shows the results on different kind of datasets.

Finally, section 6 presents the conclusions and proposes some future work.

## 2 CNNs and Faster R-CNN overview

### 2.1 CNNs

CNNs [41] [42] are a type of feed forward network able to capture local "spatial" patterns in data, exploiting spatial or temporal correlations especially in images and videos. Thus, CNNs have a special architecture that reduces the number of parameters to learn and improves the performance of back-propagation training algorithms. Generally, CNNs deal with raw data as input with minimal or no pre-processing. The core strategy consists in the derivation and application of filters to obtain suitable features. Initial layers provide primitive features (edges, borders) which are propagated through different layers of the network, obtaining richer representation usable for classification. The obtained features tend to be invariant to rotation, scale or shift since the receptive fields give the neuron access to basic features such as corners or oriented edges. Hand-crafted features such as LBP, SURF, HOG [43] [41] [44] are often outperformed by those derived using CNNs.

### 2.2 Faster R-CNN

Object detection, in addition to classification, implies localization of objects in the entire image. When video sequences are analysed, spatio-temporal techniques such as background subtraction, or optical flow, allows to differentiate object movements, which once located are further classified. Static images lack this kind of information, and traditionally, the strategy deploys a two-class classifier (object vs non-object) in conjunction with a sliding window search. Since the number of windows needed to account for different scales and aspect ratios could be very large, it is necessary to use methods like non-maximal suppression to reduce redundant candidates. So, object proposal algorithms have been created, with strategies like Branch & Bound [45], to limit search using calibration information [46] or grouping adjacent pixels merging them to find a blob region as in Selective Search [47] or using pre-defined windows based on objects candidates as in Spatial Pyramid Pooling [48], or Edge boxes [49] (see [50] for a complete comparison). More recently, CNN is used for classification purposes in combination with a pre-filtering strategy as in R-CNN [51], but demanding considerable time in the training process. To improve training process time, Fast R-CNN [29] was proposed, which swaps the locating strategy of detecting regions and running CNN. The CNN network produces a high resolution convolutional feature map which is extracted from the image. Region proposals are obtained from the feature map, feeding the convolutional features of these regions into fully connected layers, with a linear classifier and a bounding box linear regression module to define regions. Nevertheless, this architecture is still slow at test time. Faster R-CNN[30] improves this by combining features of a fully convolutional network to perform both region proposals and object detection. Instead of running a separate selective search algorithm, the model reuses the forward pass of the CNN (first step of classification) which defines the features of the images and at the same time are used for region proposal. The region proposal network (RPN) shares convolutional layers with the object classification network, so only one CNN is trained and region proposals are calculated almost for free. The rest of the convolutional layers are used to regress region bounds with scores for object proposal at each location. The RPN is implemented using a sliding window over the CNN feature map and, at each window, generating k potential bounding boxes and scores associated to their performance. This k represents the common aspect ratios that candidates to objects could fit, called anchor boxes. For each anchor box, the RPN outputs a bounding box and score per position in the image. This model speeds up the fast RCNN results, and improves object detection performance.

Faster R-CNN was the foundation of more than 125 proposed entries in ImageNet detection and localization at ILSVRC 2016 [52] and in the COCO challenge 2015 [30]. Figure 1 shows the general architecture of the Faster R-CNN. Both the region proposal network and the object classifier share fully convolutional layers, and are trained jointly. The region proposal network behaves as an attention director, determining the optimal bounding boxes across a wide range of scales and using nine candidate aspect ratios (anchor boxes) to be evaluated for object classification. In other words, the RPN tells the unified network where to look.

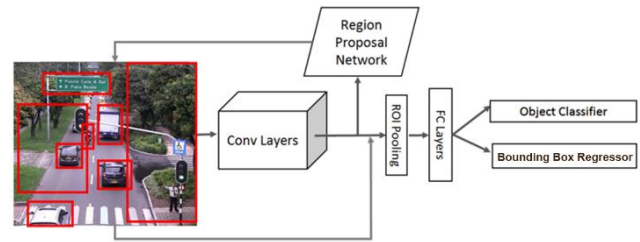

Figure 1 Faster R-CNN network structure. Modified from [33]

## 3 The Motorbike Urban Dataset

Occluded scenarios are frequent on urban traffic analysis (Figure *3*Figure 2). Vehicle detection, under this condition has been studied by many authors, benchmarking their results mainly using the KITTI dataset [53] which unfortunately lacks a motorcycle category.

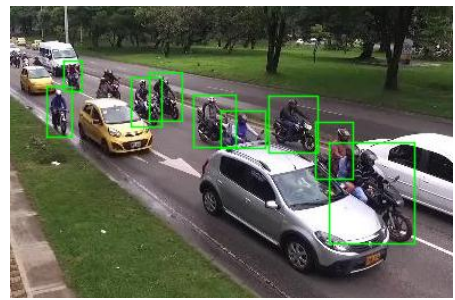

Figure 2 Sample annotated image. Note some relatively small object size and the occlusions between motorcycles and other vehicles

For this reason, we created a set of 7,500 annotated images, which includes 220 motorcycles on urban traffic. Images were taken with a Phantom 4® drone, with an HD camera under windy conditions, which affected the image stabilizer capabilities. Images were resized to 640 x 364 pixels, containing 41,040 ROI annotated objects, with a minimal height size set to 25 pixels. 60% of the annotated data corresponds to occluded motorcycles. Objects partially occluded with height less than 25 pixels were not annotated.

## 4 CNN model inspired on Faster R-CNN

The proposed model is inspired on Faster R-CNN. The network is designed using Matlab and is based on [54]. The difference here is that we create a model to deal with occlusions. A single CNN architecture has to be able to detect objects and at the same time classify them as motorcycles or not motorcycles. The region proposal mechanism uses the same CNN, so this architecture makes region proposals as part of the CNN training and prediction steps.

When the CNN is only used for classification, the input size is typically the size of the training images, but since in this model it is necessary to implement detection, smaller sections of the image have to be analysed. The input size must be similar in size to the smallest object in the data set. In this dataset the minimal annotation height size is 25, so the input size is defined as [32 32 3], to leave some pixels around the object. The input layer also implements zero centre normalization, controls the gradients and to unify the learning rate in the backpropagation training process.

The CNN model has two blocks of convolutional layers, followed by ReLU (rectified linear units), and pooling layers. The first convolutional layer includes 64 filters of [3 3], which deal with the three image channels and capture the basic primitive features. The second convolutional layer incorporates 32 filters structuring high-level image features. These high-level features are used on the recognition task, since they have a richer image representation [55]. This two-layer configuration is also used for the region proposal network (RPN). This is followed by a max ROI pooling layer with a grid of 15 x 15 pixels which can cover the minimal size of the detected objects. This layer reduces the spatial size of the extracted feature map and removes redundant spatial information. Traditional CNN architectures as AlexNet [41] may have more convolutional layers followed by max pooling layers, mainly oriented to cover more complexity in features. The proposed model with just two convolution layers, learns features with sufficient discriminate attributes to differentiate even occluded motorbikes from the background and other objects. It is important to avoid down-sampling of the data prematurely, keeping the number of pooling layers low, avoiding discarding image information that is useful for learning. The final layer corresponds to a fully connected (FC) layer of 64 output neurons, which combines all the features learned by the previous layers identifying the larger patterns. This layer is rectified by the final ReLU. Finally, the last fully connected layer combines the features to classify the images, generating outputs to valuate if a given input corresponds to motorcycles or otherwise. Normalizing the output of the last FC layer, the softmax layer quantifies the confidence of the last classification layer, which also computes the loss. Figure 3 shows the described model.

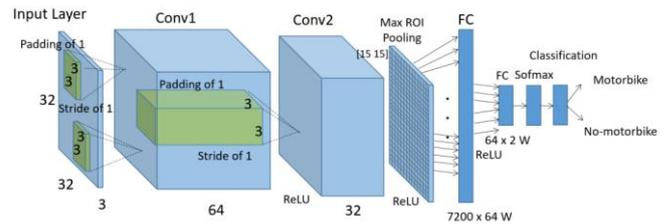

Figure 3 Proposed CNN Model.
This model is used for RPN and for classification

As in faster R-CNN, training of the model involves four steps. The first two steps train the Region Proposal Network (RPN) and the detection network, minimizing the loss and obtaining the weights and biases. The final two steps combine the results of the pre-trained networks, fusing the parameters of a single network for detection and classification. All the steps use Stochastic Gradient Descent with Momentum (SGDM) as the optimization algorithm for training.

$$\theta_{\ell+1} = \theta_\ell - \alpha \nabla E(\theta_\ell) + \gamma(\theta_\ell - \theta_{\ell-1}) \qquad (1)$$

where $\ell$ corresponds to iteration number, $\alpha > 0$ is the learning rate, $\theta$ is the parameter vector (weights and biases) and $E(\theta)$ is the loss function. The stochastic component corresponds to the evaluation of the gradient and the updates of parameters using a subset of the training set (minibatch). Each evaluation of the gradient using the mini-batch is an iteration. At each iteration, the algorithm takes one step towards minimizing the loss function. The full pass of the training algorithm over the entire training set using mini-batches is an epoch. The momentum term $\gamma$ determines the contribution of the previous gradient step to the current iteration, and is used to avoid oscillation along steepest descent to the optimum.

The learning rate of the two first steps is larger (1e-5 vs 1e-6) since the first layers require faster convergence, while the last two involve fine-tuning, as the network weights need to be modified more slowly. The number of epochs on each step is set to 60 (see section 5 for values and results). CNN algorithms can demand many epochs used for the backpropagation algorithm to converge on a combination of weights with an acceptable level of accuracy.

Especially for RPN training, images patches are extracted from the training data. Here it is important to define the positives examples and negative ones. Positive training samples are those that overlap with the ground truth boxes by 0.6 to 1.0, measured by the bounding box intersection over union metric (IoU). Negative training samples overlap by 0 to 0.3. Pyramid scaling is also used to identify motorcycles under variate sizes on the image.

To avoid overfitting, the created data set is split into training and validation data. 60% of images are used as training data (4500 images), and the remaining 40% (3000 images), for validation. The split selection is randomized to avoid biasing the results. Several tests were conducted reducing the number of examples for training and validation (see section 5 for values and results), nevertheless as it is expected in deep learning applications, the higher the examples number, the better the final AP results.

For this work we exploit the capabilities of an Nvidia Titan X (Pascal) 1531Mhz GPU, running on a Windows 10 Machine with a core i7 7$^{th}$ generation 4.7 GHz, and 32 GB of RAM. Due to the number of training examples (4500) and epochs configured, the model took 32 hours for the trained process. Different tests with less examples were also carried out (see section 5 for values and results).

## 5 Experiments and Results

We evaluate the model with three different datasets. First we use the dataset of 300 images provided by the Matlab [54] example. Although this dataset only includes cars, it is valid to evaluate the performance of the proposed model. This dataset includes a maximum of 4 annotated cars per image (Figure 4), which are not occluded by any object. We achieve 81% average precision (AP), while the example model reports only 60%.

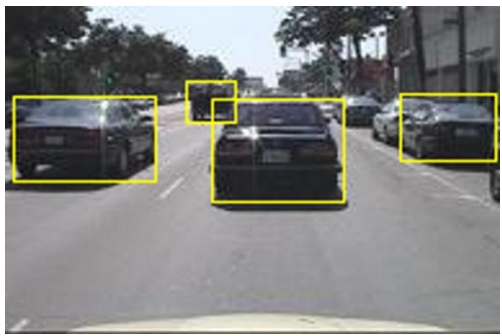

Figure 4 The MatLab dataset example, for the highest number of vehicles/image (4). Note that there is no occlusion.

The second dataset (Las Vegas) corresponds to a video sequence of 1812 RGB frames (640x480), during daylight and good weather conditions. This dataset is available upon request from the authors. The sequence annotated 36 different cars to detect (including sedan, van and taxis), 7 motorbikes and 1 Bus, which is the class assigned to a detected truck. In this dataset the proposed model achieves 92% of AP for the motorcycle class.

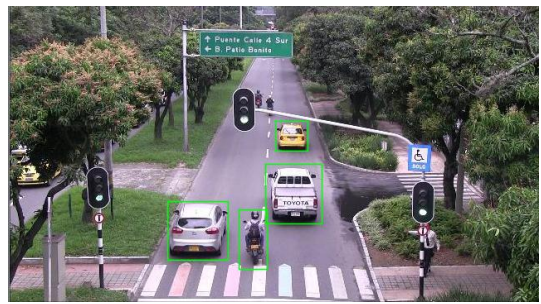

Figure 5 Las Vegas dataset.
Note the almost null occlusion of the annotated objects.

Finally, for dealing with occlusion we evaluate the challenging Motorbike Urban Dataset (also available from the authors), tuning the proposed model to obtain best result for detection and classification. The different modifications used to obtain better results are described in Table 1. It is important to note the high level of occlusion that many of the annotated objects presents. We achieve an AP of 75% (Figure 6).

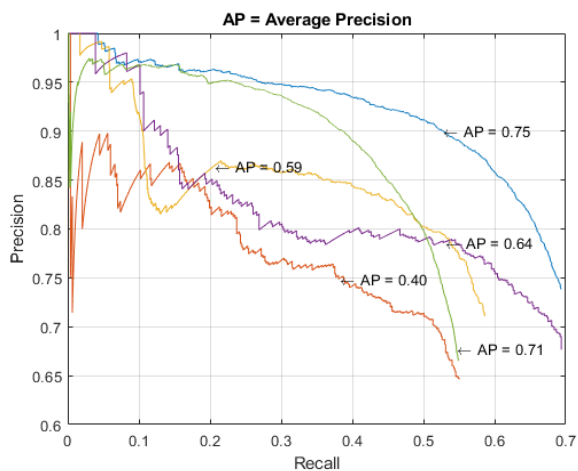

Figure 6 Average Precision (AP) of the model on the Motorbike Urban Dataset, evaluations done according to Table 2 parameters.

We compared the model against AlexNet and Faster-RCNN based on VGG16. Evaluating the dataset with AlexNet model complemented with GMM background subtraction only achieves 17% AP, while Faster R-CNN based on VGG16 raises the AP to 23%. It is important to remark that AlexNet+GMM model exploits spatio-temporal information of the video sequence using GMM background subtraction, which seems to have problems dealing with occlusions, and overlapping objects, as well with the movement of the camera.

| No Examples | Input Layer | Filter Size | Filters1 | Filters2 | FC | Epochs | LR | AP |
|---|---|---|---|---|---|---|---|---|
| 500 | 32 | [3 3] | 32 | 32 | 64 | 10 | [-5/-6] | 0.2719 |
| 500 | 25 | [3 3] | 32 | 32 | 64 | 10 | [-5/-6] | 0.3492 |
| 300 | 32 | [3 3] | 32 | 32 | 64 | 40 | [-5/-6] | 0.4083 |
| 2000 | 32 | [3 3] | 32 | 32 | 64 | 40 | [-5/-6] | 0.5023 |
| 1000 | 32 | [3 3] | 32 | 32 | 64 | 40 | [-5/-6] | 0.5962 |
| 200 | 32 | [3 3] | 32 | 32 | 64 | 40 | [-5/-6] | 0.6303 |
| 200 | 32 | [3 3] | 64 | 32 | 64 | 40 | [-5/-6] | 0.6379 |
| 3000 | 32 | [3 3] | 64 | 32 | 64 | 50 | [-5/-6] | 0.6420 |
| 200 | 32 | [3 3] | 64 | 32 | 64 | 60 | [-5/-6] | 0.7100 |
| 7500 | 32 | [3 3] | 64 | 32 | 64 | 60 | [-5/-6] | **0.7523** |

Table 1 Different configurations of the model and Average Precision results (AP) on the Motorbike Urban Dataset.

The results can be viewed via the following links https://youtu.be/qamvkieEKto - https://youtu.be/qamvkieEKto and https://youtu.be/ZJjT6fJqpUA.

## 6 Conclusions and Future Work

This paper has proposed a Faster R-CNN based model for motorcycle detection in urban scenarios. The model can deal with highly occluded images, and achieves an AP precision of 75% in a newly introduced annotated motorbike urban dataset of 7500 images. Results are comparable with state of the art algorithms published in benchmarking sites as KITTI [53].

Several tests were performed, even comparing the model performance with state of the art CNN models, which in some cases take advantage of the spatio-temporal information of the video sequences.

As most of the studied deep learning architectures, the model presents better results by using considerable examples for learning, making the training process a time-consuming task, even with the use of GPU architectures.

Future work will move toward architectures as LTRCNN [56] where models can be jointly trained to simultaneously learn temporal dynamics and convolutional perceptual representations. Detection and classification will be improved with tracking and applied to an enriched wider set of urban road user classes (e.g. trucks, vans, cyclists, pedestrians).

## Acknowledgements
S.A. Velastin is grateful to funding received from the Universidad Carlos III de Madrid, the European Union's Seventh Framework Programme for research, technological development and demonstration under grant agreement no. 600371, el Ministerio de Economía y Competitividad (COFUND2013-51509) and Banco Santander. The authors gratefully acknowledge the support of NVIDIA Corporation with the donation of the GPUs used for this research. The data and code used for this work is available upon request from the authors.